\pdfoutput=1

\documentclass[11pt]{article}

\usepackage[]{emnlp2021}

\usepackage{times}
\usepackage{latexsym}

\usepackage[T1]{fontenc}

\usepackage[utf8]{inputenc}

\usepackage{microtype}
\usepackage{graphicx}

\title{LegaLMFiT: Efficient Short Legal Text Classification with LSTM Language Model Pre-Training}

\author{Benjamin Clavi\'e \and {\bf Akshita Gheewala} \and {\bf Paul Briton} \and {\bf Marc Alphonsus} \\ \and {\bf Rym Laabiyad} \and {\bf Francesco Piccoli} \\
  Jus Mundi \\
  10 Rue de Penthièvre, Paris, France \\
  \texttt{\{firstletter.lastname\}@jusmundi.com} \\}

\begin{document}
\maketitle
\begin{abstract}
Large Transformer-based language models such as BERT have led to broad performance improvements on many NLP tasks. Domain-specific variants of these models have demonstrated excellent performance on a variety of specialised tasks.
In legal NLP, BERT-based models have led to new state-of-the-art results on multiple tasks. The exploration of these models has demonstrated the importance of capturing the specificity of the legal language and its vocabulary.
However, such approaches suffer from high computational costs, leading to a higher ecological impact and lower accessibility.
Our findings, focusing on English language legal text, show that lightweight LSTM-based Language Models are able to capture enough information from a small legal text pretraining corpus and achieve excellent performance on short legal text classification tasks. This is achieved with a significantly reduced computational overhead compared to BERT-based models. However, our method also shows degraded performance on a more complex task, multi-label classification of longer documents, highlighting the limitations of this lightweight approach.
\end{abstract}

\section{Introduction}

The recent years have seen the development of large pre-trained language models based on the Transformer \citep{transformers} architecture such as BERT \citep{bert}. These models have led to significant improvements on the existing state-of-the-art for various NLP tasks and benchmarks such as GLUE \citep{glue}.

In the wake of these impressive performances on general-domain benchmarks, subsequent work has explored the development of specialised pre-trained models. The aim of these models is to better capture the vocabulary and subtleties of various specialised domains, on which general-domain models under-performed, such as biology \citep{biobert} and medical clinical notes \citep{clinicalbert}.

Within the legal domain, similar efforts have been undertaken, showing improvements on a variety of legal NLP tasks \citep{legalbertacl}. This improvement is however less pronounced on some tasks than in other specialised domains \citep{legalberticail}, despite the widely held view that legal language is extremely specific \citep{lawlang1, lawlang2}.

The impressive performance of BERT-based models does come with trade-offs.  They are computationally expensive, requiring large volumes of data and compute resources and resulting in longer training and inference times. This creates a high barrier to entry that can result in lower accessibility, despite efforts to create lighter models \citep{distilbert, legalbertacl}. The training of such models has also been shown to carry a large financial and ecological footprint \citep{energy}, highlighting the importance of properly weighing the cost associated with them \citep{parrots}.

While not as versatile as Transformer-based models, language models based on the LSTM \cite{lstm} architecture, such as ULMFiT \citep{ulmfit}, have previously achieved state-of-the-art results on classification tasks with smaller overhead. We believe such approaches to be under-explored in specialised domains. Within the legal domain, they could be complementary to BERT-like approaches, offering strong performance on less complex tasks, such as short text classification, while being considerably more computationally efficient.

In this paper, we explore the application of LSTM-based smaller pre-trained language models to English-language text classification within the legal domain. We demonstrate that such an approach vastly outperforms other LSTM-based classifiers, showing the ability of the language model pretraining to capture additional information about the nature of legal language. We also show that these models can reach performance competitive with the current state-of-the-art on multiple classification tasks with a much lighter computational footprint at both training and inference time. \\ 
To support legal NLP research, we release LegaLMFiT, the model used to achieve these results. \\
Finally, we also highlight that our approach has shortcomings when it comes to a more complex multi-label classification task on longer text, and propose potential explanations and improvements.

\section{Related Work}
Previous work on transfer learning and domain-adaptation in NLP has largely focused on models based on BERT \citep{bert} and other large pre-trained transformer models such as GPT-3 \citep{gpt3}. Models such as BioBERT \citep{biobert} and SciBERT \citep{scibert} have greatly improved the state-of-the-art of NLP tasks related to biomedical and scientific text.

Before any domain adaptation was pursued, BERT was already shown to achieve state-of-the-art performance on multiple legal NLP tasks \citep{eur57, chalk2}. Following the trend observed in other domains, work on domain-adaptation for legal NLP then focused on creating BERT-based models specifically targeting legal text. This work has also highlighted shortcoming found in common domain-adaptation approaches, with \citet{legalbertacl} introducing Legal-BERT and specifically addressing multiple issues by comparing different domain-adaptation methods and fine-tuning parameters as well as exploring the potential of smaller models. Their approach yields new state-of-the-art results on multiple tasks, yet the performance increase remains more modest than the one observed in the biomedical domain.


\citet{legalberticail} studies possible reasons as to why the performance gains found in other domains have not been fully replicated. They introduce another Legal-BERT and experiment with multiple ways of training BERT-based models. They evaluate their models on a different set of legal NLP tasks while attempting to quantify the domain specifity of each task. They find that domain adaptation does yield substantial gains on highly domain-specific tasks. Furthermore, their results show that on such tasks, using a vocabulary obtained from legal documents produces considerably better results than a general domain vocabulary.

Outside of the legal domain, ULMFiT, introduced by \citep{ulmfit} introduced a method to pre-train an LSTM-based language model and leverage the information it learns about language to greatly improve performance and reduce data needs for downstream classification tasks. This approach, anterior to the development of BERT, yielded results that were at the time state-of-the-art on multiple text classification benchmarks. While BERT-based models have later produced better results, it is notable that the ULMFiT approach uses only WikiText-103 \citep{wiki103}, a subset of Wikipedia that contains just over 100 million words. This is a considerably smaller pre-training set than BERT, which uses a corpus of 3,300 million words \citep{bert}.

\citet{eisenschlos-etal-2019-multifit} further explore this approach and obtained then state-of-the-art performances on multiple multilingual classification tasks while using considerably less data and compute than BERT-based approaches.

Despite the strong performance of models based on LSTM language models on text classification downstream tasks, the versatility and good results of BERT-based models have led to these learning approaches being under-studied for specialised domains. This is especially true for the legal domain, which we aim to address with this paper.

\begin{table*}
\centering
\begin{tabular}{lcccc}
\hline
\textbf{Task} & \textbf{Labels \#} & \textbf{Median token \#} & \textbf{Examples \#} \\
\hline
Terms of Service \citep{claudette} & 2 & 25 & 9414 \\
Overruling \citep{legalberticail} & 2 & 24 & 2400 \\
Reasoning of the Tribunal (Arbitration) & 2 & 80 & 156525 \\
EURLEX57k \citep{eur57} & 746 & 383 & 50650\\
\hline
\end{tabular}
\caption{\label{tasks}
Information about the make-up of the four downstream classification tasks.}
\end{table*}
\section{LegaLMFiT}

Our approach is largely based on the approach followed by \citet{ulmfit} when training ULMFiT. Broadly, we follow three key steps:

\textbf{Language Model Pre-Training:} The model is initially trained as a LSTM-based Language Model, built upon a 3-layer AWD-LSTM \cite{AWDLSTM} architecture. The aim of this step is for the model to capture information about legal language and vocabulary from a large volume of documents.

\textbf{Target Data Language Model Fine-Tuning:} The language model is then fine-tuned on the data that will be used for the classification task, in order for the model to capture a better representation of the language it uses. As per \citet{ulmfit}, we use discriminative fine-tuning, where each layer is fine-tuned with a fraction of the defined learning rate as different layers capture different types of information \citep{yosinski}.

\textbf{Classifier fine-tuning:} Finally, we add classification linear layers to the language model and fine-tune the model a final time on a given classification task. This fine-tuning is gradual, all the layers of the language model starting with frozen weight and being gradually unfrozen at a rate of one per epoch until they are all fine-tuned at once.

\section{Experimental Setup}
\begin{table*}
\centering
\begin{tabular}{lccc|c}
\hline
\textbf{Model} & {\vtop{\hbox{\strut \textbf{ToS}}\hbox{\strut \textit{\centering{Pos F1}}}}} & {\vtop{\hbox{\strut \textbf{Overruling}}\hbox{\strut \textit{\centering{Mean F1}}}}} & {\vtop{\hbox{\strut \textbf{RotT}}\hbox{\strut \textit{\centering{Pos F1}}}}} & {\vtop{\hbox{\strut \textbf{EURLEX57k}}\hbox{\strut \textit{\centering{nDCG@5}}}}} \\
\hline
LegaLMFiT (Ours) & \textbf{0.788} & 0.955 & \textbf{0.836} & 71.9 \\
BiLSTM Baseline \citep{legalberticail} & 0.712 & 0.910 & N/A & N/A \\
BERT-base & 0.722 & 0.958 & 0.778 & 82.9 \\
Custom Legal-BERT \citep{legalberticail} & 0.787 & \textbf{0.974} & 0.821 & N/A \\
Legal-BERT-SC \citep{legalbertacl} & N/A & N/A & 0.818 & \textbf{84.7}\\
Legal-BERT-Small \citep{legalbertacl} & N/A & N/A & 0.798 & 84.5\\

\hline
\end{tabular}
\caption{\label{results}
Test performance of the various models. Best result in bold. Results from models other than LegaLMFiT on tasks other than \textit{RotT} are from the litterature. Metrics reported follow the relevant literature. }
\end{table*}

\begin{table*}
\centering
\begin{tabular}{lcccc|c}
\hline
\textbf{Model} & \textbf{Pretraining} & {\vtop{\hbox{\strut \textbf{Finetuning}}\hbox{\strut \textit{\centering{hours}}}}} & {\vtop{\hbox{\strut \textbf{Classifier}}\hbox{\strut \textit{\centering{hours}}}}} & {\vtop{\hbox{\strut \textbf{Inference  (Test)}}\hbox{\strut\textit{\centering{minutes}}}}}& \ \vtop{\hbox{\strut \textbf{LM Corpus}}\hbox{\strut \textit{GB}}} \\
\hline
LegaLMFiT (Ours) & \textbf{14 hours}  & \textbf{1:25} & \textbf{0:22} & \textbf{0:40} & \textbf{0.85} \\
Custom Legal-BERT & N/A (Weeks)  & 12:02 & 3:05 & 5:41 & 37 \\
Legal-BERT-SC & N/A (Weeks) & 12:02 & 3:05 & 5:41 & 11.5 \\
Legal-BERT-Small & N/A (Weeks) & 3:25 & 1:40 & 3:22 & 11.5 \\

\hline
\end{tabular}
\caption{\label{cost}
Time to pre-train, fine-tune, train and run on \textit{Reasoning of the Tribunal} as well as pre-training LM dataset size for selected models. All steps performed on an NVidia V100 16GB GPU, except for the pre-training of the BERT-based models, which was performed by others on Cloud TPUs and for which no exact duration is available.}
\end{table*}

\textbf{Pre-training Data}: The Language Model is trained using 850MB of English international legal text from four sources: \textbf{International arbitration awards} harvested from multiple arbitral tribunals comprising 35\% of the corpus. \textbf{EU legislation} documents, decisions from the \textbf{European Court of Human Rights} (ECHR) and decisions from the \textbf{European Court of Justice} (ECJ) in equal proportions complete the make-up of the corpus.  \\
\\
\textbf{Text Classification Tasks:} We evaluate the model on four downstream legal text classification tasks. Three of them are from the existing literature and a fourth one is a novel text classification task. An overview of the tasks is presented in Table \ref{tasks}.

\textbf{Existing Tasks}: \textbf{Terms of Service} \citep{claudette} and \textbf{Overruling} \citep{legalberticail} are binary classification tasks on short legal documents. EURLEX57 (Frequent) \citep{eur57} is a multi-label classification on longer documents. These tasks are  detailed in Appendix \ref{sec:downstream}.

\textbf{Reasoning of the Tribunal (\textit{RotT})}: This new task is a binary classification task using data from Investor-State Dispute Settlements (ISDS) arbitration awards. Investment and trade treaties can contain an ISDS clause to resolve disputes between private investors and states before an arbitral tribunal \citep{ISDS}. It is an increasingly important part of public international law \citep{ISDSgrowing}. ISDS arbitration awards are the decisions of those arbitral tribunals. The aim of the \textit{RotT} classification task is to identify whether a given paragraph from an arbitration award is the expression of the tribunal's legal reasoning or not. The data was gathered from multiple arbitral tribunals and labeled by a team of legal experts. \\
\\
\textbf{Pre-Processing and Tokenization:} We adopt a very basic pre-processing step of replacing all digits with the digit \textbf{0}. We then perform subword tokenization using the unigram language model through SentencePiece \citep{token} and a vocabulary size of 32,000. \\
\\
\textbf{Model training} We do not perform an extensive hyperparameter search. The model is trained following the training procedures outlined by \citet{ulmfit} and using the ranger \cite{ranger} optimiser. Both the \textbf{initial language model training} and the \textbf{language model fine-tuning} are performed for ten epochs with a learning rate of 0.001, a batch size of 128 and a dropout probability of 0.3. The final classifier fine-tuning is performed for 5 epochs with a learning rate of 0.05. \\
\\
\textbf{Baselines} For the Reasoning of the Tribunal task, we train baseline models using BERT-base as well as with Legal-BERT introduced by \citet{legalbertacl} and Custom Legal-BERT from \citet{legalberticail}. In the case of Legal-BERT, we use both Legal-BERT-SC, a model trained from scratch, and Legal-BERT-Small, a smaller version of the model with strong performance but a noticeably smaller footprint.  For all BERT-based models, we first fine-tune them on the target dataset before training the classifier. For all other tasks, we use results reported in the relevant literature as a baseline.

\section{Results and Discussion}
\textbf{Short Legal Text Classification Results} The first three columns of Table \ref{results} present the results of the various classifiers on short text classification tasks. In two out of the three tasks, LegaLMFiT reaches the highest performance, narrowly reaching state-of-the-art performance on the \textit{Terms of Service} dataset. On the \textit{Reasoning of the Tribunal} dataset, we notice a significantly better performance from all the legal domain specific models over BERT-base, possibly indicating the use of more specialised language. It is also on this dataset that LegaLMFiT most significantly outperforms other models, potentially reflecting the high proportion of arbitration awards used during the pretraining of its language model.
On the \textit{Overruling} dataset, LegaLMFiT is outperformed by both Custom Legal-BERT and BERT-Base, although it comes close to BERT-Base's performance and reaches noticeably better results than a non-language model BiLSTM baseline. As in the previous case, a potential explanation for this could be found in the pre-training dataset, as it does not contain US court rulings while this task comprises entirely of extracts from US court rulings.

\textbf{EURLEX57k Classification Results} LegaLMFiT performs very poorly on the EURLEX57K multi-label classification task, and is vastly outperformed by both BERT-Base and Legal-BERT. This highlights the limitations of our approach, which appears badly suited to this task. Possible explanations reside in the the additional complexity of a multi-label task and the substantially longer content of each example in this dataset. This could partially stem from the lack of the Transformer's attention mechanism and its ability to better capture key information \citep{transformers} and which has been shown to increase the performance of LSTM-based language models \citep{sharnn}.

\textbf{Efficiency Comparison} Table \ref{cost} presents the training time on \textit{Reasoning of the Tribunal} as well as the pre-training data volume used by each model. LegaLMFiT requires considerably less resources, both in terms of training data and compute, than BERT-based approach despite beating all of them in this task. This efficiency allows the model to be considerably easier to train and deploy with a lower environmental footprint.

\section{Conclusions and Future Work}

We release LegaLMFiT, an LSTM-based language model based on the ULMFiT architecture. We showed that this model is able to reach strong performance on multiple short legal text classification tasks, even matching or exceeding the performance of current state-of-the-art methods. We also highlighted that these models require substantially less pre-training data and compute than Transformer-based approaches. Our results also show that our method performs badly on a more complex multi-label classification task with longer documents, suggesting it is not without tradeoffs. Despite this, the lightweight approach and good results obtained by LegaLMFiT make it a valuable new tool, especially in low resource settings. In future work, we plan to further research the potential of non-Transformer models within the legal domain by exploring structural improvements, such as the addition of simplified attention mechanism \citep{sharnn}. We also plan to explore how this approach performs on non-English languages.

\section{Acknowledgements}
This work was granted access to the HPC resources of IDRIS under the allocation AD011012667 made by GENCI.

\bibliography{anthology,custom}
\bibliographystyle{acl_natbib}

\appendix



\section{Existing Downstream Classification Tasks}
\label{sec:downstream}
\textbf{Terms of Service (ToS)}: Introduced by \citet{claudette}. This is a binary classification task aiming to identify whether a given clause found in online consumer contracts (terms of service) is likely to be unfair to the consumer, as per the definition of the European Union \citep{EUunfair}. The data was annotated by lawyers into three categories, \textit{clearly fair}, \textit{potentially unfair} and \textit{clearly unfair}. Following \citet{legalberticail}, to dichotomise this task, \textit{clearly fair} clauses constitute negative examples while the latter two are both examples of potentially unfair clauses.

\textbf{Overruling}: Introduced by \citet{legalberticail}. It is a binary classification task whose aim is identifying whether or not a given sentence in a legal decision contains a statement nullifying a previous decision, thus creating a new precedent replacing the previous one. The dataset was provided by Casetext and contains sentences from U.S court decisions with overruling ones annotated as such by human lawyers. 

\textbf{EURLEX57k:} Introduced by \citet{eur57}, this is a multi-label classification task of various EU laws into categories they belong to. We only attempt to classify the \textit{Frequent} labels, meaning labels which appear on 50 or more training documents.
\end{document}